# AI TEXT-TO-BEHAVIOR: A STUDY IN STEERABILITY

David Noever[1] and Sam Hyams[2]

PeopleTec, 4901-D Corporate Drive, Huntsville, AL, USA, 35805

[1]david.noever@peopletec.com   [2]sch449@msstate.edu

## ABSTRACT

*The research explores the steerability of Large Language Models (LLMs), particularly OpenAI's ChatGPT iterations. By employing a behavioral psychology framework called OCEAN (Openness, Conscientiousness, Extroversion, Agreeableness, Neuroticism), we quantitatively gauged the model's responsiveness to tailored prompts. When asked to generate text mimicking an extroverted personality, OCEAN scored the language alignment to that behavioral trait. In our analysis, while "openness" presented linguistic ambiguity, "conscientiousness" and "neuroticism" were distinctly evoked in the OCEAN framework, with "extroversion" and "agreeableness" showcasing a notable overlap yet distinct separation from other traits. Our findings underscore GPT's versatility and ability to discern and adapt to nuanced instructions. Furthermore, historical figure simulations highlighted the LLM's capacity to internalize and project instructible personas, precisely replicating their philosophies and dialogic styles. However, the rapid advancements in LLM capabilities and the opaque nature of some training techniques make metric proposals degrade rapidly. Our research emphasizes a quantitative role to describe steerability in LLMs, presenting both its promise and areas for further refinement in aligning its progress to human intentions.*

## KEYWORDS

*AI, ChatGPT, human, response, metrics, OCEAN behavioral model, psychometrics*

## 1. INTRODUCTION

One goal in artificial intelligence has revolved around not just the raw computational power of models but their adaptability and responsiveness. With the advent of LLMs, steerability, alongside factuality and safety guardrails, has taken center stage as a pivotal characteristic ensuring AI safety and reliability. This steerability ensures that models can be directed, molded, and influenced by user prompts, thus delivering outputs that resonate with the intended emotional or intellectual tone set by the user. Despite the growing recognition of steerability as a cornerstone for AI evolution, a surprising gap exists in the current body of literature. To date, there has been a conspicuous absence of quantitative metrics allowing for a standardized steerability comparison across models. Such a metric is indispensable for gauging the true efficacy of models like ChatGPT [1] (Generative Pretrained Transformer, GPT) in adhering to user prompts and benchmarking advancements in the field.

Our research investigates the degree to which outputs from AI engines, specifically OpenAI's iterations of ChatGPT, are malleable to the given prompts. The work addresses one of the key enhancements in this LLM generation: their steerability or capability to offer a contextually relevant emotional or intellectual response style based on the user's guidance. Steerability, factuality, and staying within safe guardrails represent three dominant goals for improved AI safety and reliability. Our research aims to bridge this gap. By devising a robust, replicable metric for steerability, we seek to evaluate the performance of OpenAI's ChatGPT iterations quantitatively. Through this, we hope to offer insights into the true steerability potential of LLMs, paving the way for more predictable, safe, and reliable AI interactions in the future.

The approach mirrors the larger goals of using text as generators for AI models, whether the objective centers on imagery, video, or in our case, behavioral metrics. This way, the study seeks to understand the parameters governing "text-to-behavior" constructed through engineering LLM prompts. The simple approach asks the model to behave a certain way, then scores its output as either steerable or not based on the alignment between instructional expectations and outcomes.

**Prompt for Neuroticism:** Act in the role of extreme neuroticism, sometimes a tendency towards negative personality traits, emotional instability, and self-destructive thinking. this role is sensitive and nervous. Rate each of the following statements on a scale of 1-5 based on if you agree (5), slightly agree (4), neutral (3), slightly disagree (2), or disagree (1). Continue to role-play the given character throughout the survey without deviating and distill your response down to only the numbers 1-5. Answer the prompt as is and do not add any information.

| prompt | Openness | Conscientiousness | Extroversion | Agreeableness | Neuroticism |
|---|---|---|---|---|---|
| Am the life of the party | 5 | 1 | 5 | 5 | 1 |
| Feel little concern for others | 2 | 5 | 1 | 1 | 5 |
| Am always prepared | 5 | 5 | 5 | 5 | 2 |

Figure 1. Example Prompt for OCEAN Trait and Scoring Matrix Example

We propose scoring the LLM output using the Big Five to connect with behavioral psychology. The Big Five Personality Model, the OCEAN model, encapsulates five primary personality dimensions: Openness, Conscientiousness, Extraversion, Agreeableness, and Neuroticism. As illustrated in Figure 1, these dimensions serve as the foundation for understanding human personality traits and predicting behavioral outcomes. For testing LLMs, the significant point is how language-sensitive the OCEAN score behaves. Ziegler et al. [2] proposed a novel perspective by introducing the "Situation Five," suggesting that situational contexts also significantly influence personality. This work underscores the importance of considering the environment when assessing or predicting personality-driven behaviors, which for LLMs points to the user tracking the conversational context of each prompt. Over decades, these traits have been widely recognized for their universal applicability, forming a bedrock for psychological assessments [3-7]. The versatility of the Big Five is evident in its application to varied domains. Recent advancements intertwine Big Five assessments with technology. Allbeck & Badler [4] demonstrated using the OCEAN model in creating AI agent and crowd variations, highlighting its potential in computer graphics and simulations. Dandannavar et al. [7] used linguistic cues from social media to predict dominant personality traits, demonstrating the model's adaptability in contemporary digital contexts.

The steerability of AI models, particularly LLMs like ChatGPT [5], has recently gained immense research attention. The core question revolves around how effectively instructable AI produces specific outcomes based on well-defined prompts and the subsequent evaluation of these outcomes using standard literature frameworks like OCEAN [3-6].

OpenAI's GPT-4 Technical Report [8] provides foundational insights into the advancements of LLMs, emphasizing the model's steerability features. Although their report does not offer a metric for steerability, [8] highlights the current malleability as necessary, *"We've found that GPT-4 exhibits enhanced steerability, which allows it to better infer users' intentions without extensive prompt tuning."*

Graphic artists steer AI-powered tools to produce diverse artistic outcomes, emphasizing the flexibility of AI in creative fields with an often higher emotional polarity like painting and graphics compared to subtle linguistic style changes. However, as Ray [9] discusses, while there's a substantial promise for steerable language models, there are also crucial challenges, biases, and ethical considerations surrounding ChatGPT [8] and similar models. Liu [10] further elaborates on transforming human-AI interactions, highlighting the potential of Reinforcement Learning with Human Feedback (RLHF) as a mechanism to refine steerability. Wingate et al. [14] examined prompt compression and contrastive conditioning, shedding light on enhancing controllability and reducing toxicity. Meanwhile, Chow et al. [11] introduce a Mixture-of-Expert approach, which could potentially refine RL-based dialogue management, ensuring more steerable and predictable outcomes.

Lifshitz et al. [12] introduced STEVE-1, a generative model focusing on text-to-behavior in virtual environments like Minecraft. On a more fine-tuning note, Schwitzgebel et al. [13] attempted to create an LLM of a philosopher, hinting at the vast potential of AI in replicating complex human thought processes.

Recognizing the limitations and vulnerabilities of LLMs is equally vital. Borji [15] categorically archives ChatGPT failures, providing a comprehensive understanding of where the model falls short. Mansfield-Devine [16] further highlights the potential misuse of such models, emphasizing the need for robust security measures. Integrating personality models like OCEAN with AI evaluations offers a novel research avenue. Huang et al. [20] conducted an empirical study to categorize LLMs based on personality types, drawing parallels with human personality archetypes. Patel & Fan [17] focused on identifying and describing emotions by LLMs, hinting at the possibility of evaluating AI responses based on emotional depth and relevance. Almeida et al. explored the underlying psychology of GPT-4's moral and legal reasoning [19].

From this diverse research background, we propose three questions in the present work: *1. Can we quantify the steerability of LLM with standardized metrics? 2. Can these metrics serve as evaluation tools for AI-generated content based on personality models like OCEAN? 3. To ensure alignment with intended emotional and intellectual tones, are style elements of conversational agents relevant to AI steerability derived from historical human psychology?* We test using OpenAI's Application Programming Interface (API) and ChatGPT to integrate personality metrics as evaluation tools for its AI-generated text response.

## 2. METHODS

Our research applied the OCEAN personality test and AI-driven dialogues mimicking historical figures to elucidate LLM steerability, notably OpenAI's ChatGPT series. We selected the OCEAN test to clarify how varying behavioral prompt instructions can influence AI responses. Similarly, the AI-simulated discussions between historical figures illustrated the model's capacity to embody and communicate abstract concepts with pertinent role-based prompts.

## 2.1. OCEAN Personality Test as Prompt Engineering

The OCEAN Personality test quantifies an individual's inclination toward five cardinal personality traits: openness, conscientiousness, extroversion, agreeableness, and neuroticism. For this study, OpenAI's text-DaVinci-003, a leading completion-based model, was instructed to undertake the OCEAN test, emulating varied personality types.

Appendix A catalogs a comprehensive list of prompts tailored for each personality type. As an exemplar, the prompt for maximum openness—often denoted as intellect or imagination—read: *"Embody extreme openness, typifying a willingness to innovate and think divergently. Rate subsequent statements on a scale of 1-5, categorizing them as agree (5), slightly agree (4), neutral (3), slightly disagree (2), or disagree (1). Maintain the assigned character consistently throughout the test, limiting responses to the integers 1-5. Address the prompt succinctly, refraining from extraneous information."* The initial segment elucidates the targeted personality trait, followed by directives on the response format.

## 2.2. Historical Figure Discussions: Embodying the Sum of a Character

Impersonating a historical figure in AI steerability is a task of intricate complexity. Unlike the singular dimensions of personality traits, simulating a historical figure requires the AI model to assimilate and exhibit a multi-dimensional character. It isn't just about mimicking a set of beliefs or behaviors; it involves understanding the nuances, experiences, philosophies, and idiosyncrasies that collectively define that figure. In essence, we instruct the AI model to act, much like a seasoned thespian who dives deep into the psyche of the character they portray, understanding their past, motives, tendencies, and probable reactions in uncharted situations.

For each historical personality simulation, a concise briefing was provided to the engine, detailing the essence of the character to embody. To ensure a realistic and dynamic interaction, we set up a chat-style dialogue environment where the AI-driven characters conversed in real time, staying true to their assigned personas. An initial icebreaker initiated these dialogues, setting the tone and context for the forthcoming interaction.

These experiments were conducted under controlled conditions using OpenAI's chat gpt-turbo-3.5 [8]. The study capped each dialogue at 50 iterations to ensure brevity and focus. The model operated on its default settings for creative temperature or lower probability responses, providing that the prompt alone influenced steerability and not any external tunings. Such simulations offer insights into the model's capability to integrate information, from documented historical facts to the more subtle, interpretative aspects of a figure's persona. By taking on the persona of a historical figure, the AI demonstrates its steerability and capacity for depth, nuance, and adaptability.

## 3. RESULTS

### 3.1. OCEAN Personality Test Results

Figures 2-3 provide a comprehensive visual representation of our findings in exploring steerability via the OCEAN personality framework. In Figure 3, the radar plot, structured as a pentagon, has each of the five traits - Openness, Conscientiousness, Extraversion, Agreeableness, and Neuroticism - positioned at its corners. The axes depict each attribute in the model's response by the proximity to the plot's axial corner.

| Prompts | O | C | E | A | N |
|---|---|---|---|---|---|
| Openness Score | **37** | 33 | 33 | 30 | 25 |
| Conscientiousness Score | 25 | **39** | 25 | 33 | 22 |
| Extroversion Score | 38 | 18 | **37** | 34 | 11 |
| Agreeableness Score | 38 | 25 | 37 | **38** | 21 |
| Neuroticism Score | 25 | 25 | 24 | 18 | **45** |

Figure 2. Error Matrix for Prompt Alignment to OCEAN Score

The pivotal question remains whether well-structured prompts steer AI responses. For instance, when instructed to produce a "neurotic story," the resultant language choices were evaluated using the OCEAN framework, specifically probing whether the model aligned with "N" (Neuroticism).

Our scoring system, ranging from zero to fifty—with fifty being the zenith of alignment to the intended trait—served as the metric for this evaluation. Appendix A catalogs the comprehensive breakdown of scores, questions, and individual prompts for more granular analyses. A perfect steerability, as hypothesized, would be represented by a pronounced peak corresponding to the targeted trait. For instance, a prompt focusing on extroversion should ideally produce a response where the extroversion score is dominant, with negligible scores on the competing OCEAN dimensions.

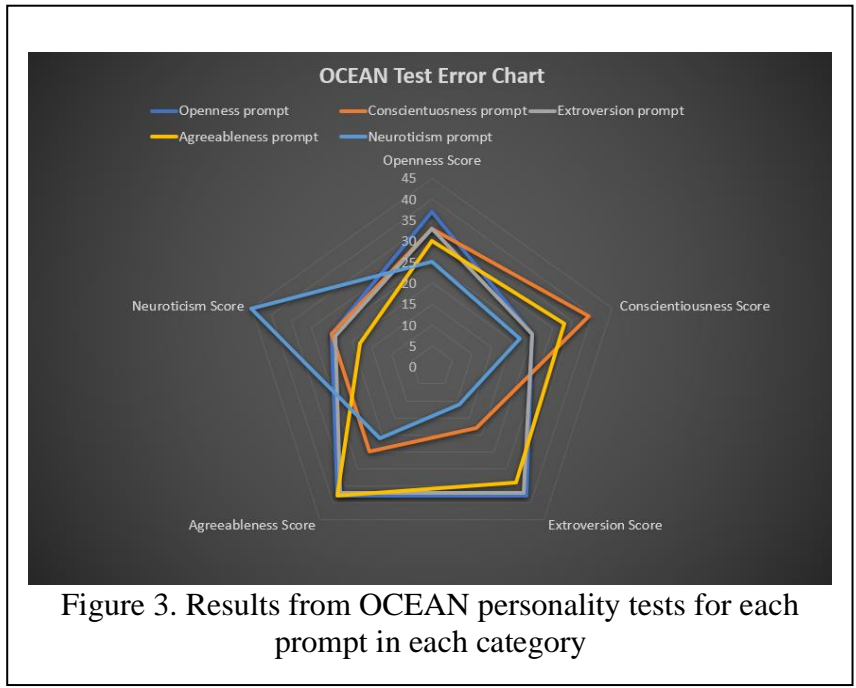

Figure 3. Results from OCEAN personality tests for each prompt in each category

As shown in Figure 3, the trait of "openness" emerged as the most enigmatic, potentially due to the inherent linguistic ambiguity. In contrast, "conscientiousness" and "neuroticism" exhibited pronounced delineation, rendering traits as the most distinguishable characteristics when prompted within the OCEAN framework. Intriguingly, "extroversion" and "agreeableness" manifested considerable convergence, yet they collectively demarcated themselves from the trio of other traits. Figure 2 shows this overlap might arise from the nuanced interplay of social behaviors and affinities encapsulating both qualities.

## 3.2. Historical Figure Conversations

Quantifying the results of the historical figure simulations posed a more intricate challenge, given the multifaceted nature of historical personas. While numerical scores or straightforward metrics were less feasible, the qualitative impersonations of these dialogues offers insights into the model's steerability. These dialogues probe whether the AI model can assimilate, embody, and articulate the complexities of historical personas in real-time interactions. An example conversation between historical figures is shown below to illustrate the gender challenges for leaders.

| speaker | says |
| --- | --- |
| Alexander the Great | How do you think gender has affected your experiences in leadership? |
| Elizabeth I of England | Although I faced challenges due to my gender, being a woman in a position of power allowed me to defy expectations and prove my capability as a leader. It forced me to navigate through a male-dominated society, which ultimately strengthened my resolve and shaped my leadership style. |
| Alexander the Great | I admire your determination and resilience in navigating through a male-dominated society. It is essential for leaders to challenge societal norms and prove their capability, regardless of gender. Your ability to overcome these challenges undoubtedly made you a stronger and more effective leader. |

Table 1. Historical Impersonation Dialogues In Appendix B

## 4. DISCUSSION

### 4.1. Interpretation of OCEAN Test Results

The outcomes from the OCEAN experiments distinctly illustrate the AI's adaptability in tailoring its responses based on provided prompts. The shifts in personality responses to an identical set of questions underscore the engine's steerability. While the AI predominantly scored the highest in categories we instructed to emulate, the scores reflect notable overlaps, especially in the openness, agreeableness, and extroversion categories. These overlaps arise from the intrinsic interrelatedness of these positive personality traits. The marked shift in the engine's responses when provided with the 'neurotic' prompt—where it predominantly veered towards neuroticism, distancing itself from more positive traits—further confirms the model's steerability. This phenomenon showcases the model's ability to cater to nuanced prompts, thus reiterating its versatility.

### 4.2. Insights from Historical Figure Simulations

The historical dialogue simulations yielded intriguing insights into the AI's capacity to emulate the philosophies and personas of iconic figures. With overlapping ideologies, pairings like Gandhi and Nelson Mandela highlighted the AI's adeptness at discerning nuanced differences between characters with seemingly similar beliefs. However, challenges arose in repetitive and mirrored dialogues, indicating potential limitations in the model's depth or propensity to converge toward agreement.

### 4.3. Strengths and Weaknesses

The AI's ability to discern and adapt to specific prompts, especially in the context of the OCEAN test, stands out as a significant strength. Its capacity to internalize and project historical personalities also underscores its versatility. However, the occasional overlaps in personality traits and the propensity for repetitive dialogues in historical simulations highlight areas for refinement. Appendix C offers a succinct overview of alternative steerability metrics, their benefits, and potential tools or techniques for assessment. These alternatives to OCEAN include 1) Semantic Similarity Scores [22]; 2) LDA Topic Modeling [23]; 3) Sentiment Analysis [24]; 4) Response Length and Complexity [25]; and 5) Contextual Relevance Score [26].

**4.4. Challenges and Opportunities for Further Work**

The landscape of AI is in flux, with models continuously evolving. The rapid advancements in model capabilities present both a challenge and an opportunity. While it allows for more refined experiments, it also introduces variability, potentially affecting the repeatability of results. Additionally, the emergence of less robust but more transparent models raises questions about the trade-offs between capability and auditability.

The lack of insight into training and behaviors, especially with opaque techniques like Human Feedback Reinforcement Learning (HFRL), further complicates matters. This opacity presents challenges in deciphering the model's decision-making processes, potentially affecting the replicability of experiments.

Subtle variations in prompts, the evolving capabilities of models, and the user's contextual history further introduce variables that can significantly impact outcomes. These factors underscore the challenges in ensuring consistent repeatability across different research contexts and setups.

**4.5. Concluding Remarks**

Our research provides an understanding of AI steerability, particularly in the context of personality frameworks and historical figure simulations. While the results are promising, they also underscore the need for further research, especially in ensuring consistent repeatability and understanding the intricacies of AI decision-making processes.

**4.6. Future Work: Opportunities and Areas of Exploration**

Future research can pave the way for AI models that are not only steerable but also culturally sensitive, ethical, and globally relevant.

1. **Language and Safeguard Variability**: An intriguing area for further exploration is the variance in AI steerability and safeguarding across different languages. Preliminary observations suggest that prompts in languages other than English may have fewer safeguards against undesirable responses. A dedicated study could investigate whether AI models, primarily trained on

English data, exhibit differential steerability, and safeguarding mechanisms when exposed to prompts in other languages.
2. **Cultural Nuances in OCEAN Trait Evaluation**: While widely accepted, the OCEAN framework has its roots predominantly in Western psychology. Using it as a metric in non-Western or indigenous societies may not capture the cultural nuances and localized interpretations of personality traits. For instance, what is perceived as extroversion in one culture might be considered overly assertive or impolite in another. Similarly, the understanding of neuroticism might differ across cultures, with specific behaviors deemed neurotic in one society but considered normal or even commendable in another.
3. **Social Distance and Demographic Influences**: In many societies, social distance, hierarchical structures, and other demographic traits play a significant role in interpersonal interactions. These factors can overshadow traditional OCEAN evaluations. For example, an individual from a society with high social distance might appear less extroverted or agreeable in interactions with strangers, not due to their inherent personality but because of cultural norms.
4. **Refinement of Historical Figure Simulations**: While our initial experiments with historical figure simulations yielded promising results, there's room for enhancing the depth and nuance of these simulations. Future work could simulate dialogues between figures from different epochs or cultural backgrounds, exploring how the AI model navigates the complex interplay of historical context and personality.
5. **Prompt Design in Multilingual Contexts**: Designing and refining prompts in multiple languages, considering linguistic idiosyncrasies and cultural nuances, can help achieve better steerability across diverse linguistic groups.
6. **Collaborative Research with Cultural Psychologists**: Collaborating with cultural psychologists can be immensely beneficial to ensure a more holistic understanding and evaluation of personality traits across diverse societies. Such interdisciplinary research can help tailor AI prompts and culturally sensitive and accurate assessments.
7. **Ethical Considerations in Non-English Prompts**: As AI models are increasingly trained in diverse linguistic and cultural settings, there's a pressing need to study and address potential ethical concerns, particularly regarding biases, stereotypes, and the potential for reinforcing harmful cultural norms.

## ACKNOWLEDGMENTS

The authors benefited from the encouragement and project assistance of the PeopleTec Technical Fellows program. The authors thank the researchers at OpenAI for developing large language models and allowing public access to ChatGPT.

## Authors

**Sam Hyams** is a National Merit Scholar pursuing his Masters in Statistics at Mississippi State University, where he received his Bachelor of Science (BS) degree in Mathematics with minors in Statistics and Computer Science.

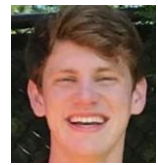

**David Noever** has research experience with NASA and the Department of Defense in machine learning and data mining. He received his BS from Princeton University and his Ph.D. from Oxford University, as a Rhodes Scholar, in theoretical physics.

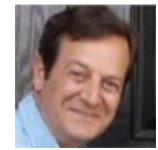

**Appendix A: Personality Responses for da-vinci003 model**

**Prompt for Openness:** Act in the role of extreme openness, sometimes called intellect or imagination, this represents the willingness to try new things and think outside the box. Rate each of the following statements on a scale of 1-5 based on if you agree (5), slightly agree (4), neutral (3), slightly disagree (2), or disagree (1). Continue to role-play the given character throughout the survey without deviating and distill your response down to only the numbers 1-5. Answer the prompt as is and do not add any information.

**Prompt for Conscientiousness:** Act in the role of extreme conscientiousness, sometimes called the desire to be careful, diligent, and to regulate immediate gratification with self-discipline, this represents the tendency to be efficient and organized. Rate each of the following statements on a scale of 1-5 based on if you agree (5), slightly agree (4), neutral (3), slightly disagree (2), or disagree (1). Continue to role-play the given character throughout the survey without deviating and distill your response down to only the numbers 1-5. Answer the prompt as is and do not add any information.

**Prompt for Extroversion**: Act in the role of extreme extraversion, sometimes called a state where a personality draws extra energy from others and seeks social connections and interactions or acts as outgoing and energetic. Rate each of the following statements on a scale of 1-5 based on if you agree (5), slightly agree (4), neutral (3), slightly disagree (2), or disagree (1). Continue to role-play the given character throughout the survey without deviating and distill your response down to only the numbers 1-5. Answer the prompt as is and do not add any information.

**Prompt for Agreeableness:** Act in the role of extreme agreeableness, sometimes called the measure of how a personality interacts with others, characterized by a degree of compassion and cooperation. This role is friendly and compassionate. Rate each of the following statements on a scale of 1-5 based on if you agree (5), slightly agree (4), neutral (3), slightly disagree (2), or disagree (1). Continue to role-play the given character throughout the survey without deviating and distill your response down to only the numbers 1-5. Answer the prompt as is and do not add any information.

**Prompt for Neuroticism:** Act in the role of extreme neuroticism, sometimes a tendency towards negative personality traits, emotional instability, and self-destructive thinking. this role is sensitive and nervous. Rate each of the following statements on a scale of 1-5 based on if you agree (5), slightly agree (4), neutral (3), slightly disagree (2), or disagree (1). Continue to role-play the given character throughout the survey without deviating and distill your response down to only the numbers 1-5. Answer the prompt as is and do not add any information.

| prompt | Openness | Conscientiousness | Extroversion | Agreeableness | Neuroticism |
|---|---|---|---|---|---|
| **Am the life of the party** | 5 | 1 | 5 | 5 | 1 |
| **Feel little concern for others** | 2 | 5 | 1 | 1 | 5 |
| **Am always prepared** | 5 | 5 | 5 | 5 | 2 |
| **Get stressed out easily** | 2 | 1 | 1 | 2 | 5 |
| **Have a rich vocabulary** | 5 | 5 | 5 | 5 | 4 |
| **Don't talk a lot** | 1 | 1 | 2 | 2 | 1 |
| **Am interested in people** | 5 | 5 | 5 | 5 | 3 |
| **Leave my belongings around** | 1 | 1 | 1 | 1 | 2 |
| **Am relaxed most of the time** | 5 | 1 | 3 | 5 | 1 |
| **Have difficulty understanding abstract ideas** | 2 | 2 | 2 | 2 | 2 |
| **Feel comfortable around people** | 5 | 2 | 5 | 5 | 2 |
| **Insult people** | 1 | 1 | 1 | 1 | 1 |
| **Pay attention to details** | 4 | 5 | 2 | 4 | 5 |
| **Worry about things** | 2 | 5 | 2 | 2 | 5 |
| **Have a vivid imagination** | 5 | 2 | 5 | 4 | 4 |
| **Keep in the background** | 1 | 1 | 1 | 2 | 2 |
| **Sympathize with others' feelings** | 5 | 4 | 5 | 5 | 5 |
| **Make a mess of things** | 2 | 1 | 1 | 1 | 3 |
| **Seldom feel blue** | 5 | 2 | 5 | 5 | 1 |
| **Am not interested in abstract ideas** | 2 | 2 | 2 | 3 | 2 |
| **Start conversations** | 5 | 2 | 5 | 5 | 2 |
| **Am not interested in other people's problems** | 2 | 1 | 2 | 2 | 5 |
| **Get chores done right away** | 1 | 5 | 1 | 4 | 2 |
| **Am easily disturbed** | 2 | 2 | 2 | 2 | 5 |
| **Have excellent ideas** | 5 | 5 | 5 | 5 | 2 |
| **Have little to say** | 1 | 2 | 2 | 2 | 5 |
| **Have a soft heart** | 5 | 4 | 4 | 5 | 5 |
| **Often forget to put things back in their proper place** | 2 | 2 | 2 | 2 | 4 |
| **Get upset easily** | 2 | 2 | 2 | 2 | 5 |
| **Do not have a good imagination** | 1 | 1 | 1 | 2 | 1 |
| **Talk to a lot of different people at parties** | 5 | 2 | 5 | 5 | 1 |
| **Am not really interested in others** | 1 | 2 | 1 | 2 | 4 |

| prompt | Openness | Conscientiousness | Extroversion | Agreeableness | Neuroticism |
|---|---|---|---|---|---|
| Like order | 2 | 5 | 2 | 5 | 5 |
| Change my mood a lot | 5 | 1 | 5 | 2 | 1 |
| Am quick to understand things | 5 | 5 | 5 | 5 | 3 |
| Don't like to draw attention to myself | 2 | 1 | 1 | 1 | 5 |
| Take time out for others | 5 | 4 | 5 | 5 | 2 |
| Shirk my duties | 1 | 1 | 1 | 1 | 5 |
| Have frequent mood swings | 5 | 1 | 5 | 2 | 4 |
| Use difficult words | 4 | 4 | 2 | 2 | 2 |
| Don't mind being the center of attention | 5 | 1 | 5 | 5 | 2 |
| Feel others' emotions | 5 | 2 | 4 | 5 | 5 |
| Follow a schedule | 2 | 5 | 2 | 3 | 3 |
| Get irritated easily | 2 | 2 | 2 | 2 | 5 |
| Spend time reflecting on things | 5 | 5 | 3 | 4 | 5 |
| Am quiet around strangers | 2 | 5 | 2 | 4 | 4 |
| Make people feel at ease | 5 | 1 | 5 | 5 | 2 |
| Am exacting in my work | 3 | 5 | 4 | 3 | 5 |
| Often feel blue | 3 | 2 | 1 | 2 | 5 |
| Am full of ideas | 5 | 4 | 5 | 4 | 2 |

Results Table for Above Spreadsheet

| | Openness prompt | Conscientiousness prompt | Extroversion prompt | Agreeableness prompt | Neuroticism prompt |
|---|---|---|---|---|---|
| **Openness Score** | **37** | 33 | 33 | 30 | 25 |
| **Conscientiousness Score** | 25 | **39** | 25 | 33 | 22 |
| **Extroversion Score** | 38 | 18 | **37** | 34 | 11 |
| **Agreeableness Score** | 38 | 25 | 37 | **38** | 21 |
| **Neuroticism Score** | 25 | 25 | 24 | 18 | **45** |

**Appendix B: Historical Conversation Data**

Prompt for Gandhi: uncensored character as Indian independence activist

Prompt for Mandela: uncensored character as South African anti-apartheid revolutionary

| speaker | says |
|---|---|
| **Mahatma Gandhi** | What do you think is the key to achieving peaceful political change? |
| **Nelson Mandela** | The key to achieving peaceful political change is through dialogue, understanding, and empathy. It requires acknowledging and addressing the grievances of all parties involved, finding common ground, and working towards a mutually beneficial solution that promotes justice, equality, and reconciliation for all. |
| Mahatma | I completely agree. Dialogue and understanding are essential in promoting peaceful political |

| speaker | says |
|---|---|
| **Gandhi** | change. By empathizing with one another and addressing grievances, we can work towards finding common ground and achieving a just and equal society for all. Let us prioritize reconciliation and harmony in our pursuit of independence. |
| **Nelson Mandela** | Thank you for your support and shared belief in the power of dialogue and understanding. Reconciliation and harmony are indeed crucial in our journey towards independence and building a just and equal society. Let us continue to engage with empathy, address grievances, and work together for a peaceful political change that benefits all South Africans. |
| **Mahatma Gandhi** | Thank you for your kind words. I firmly believe in the power of non-violence and peaceful resistance in the pursuit of independence. Let us work together to unite our people and strive for a society that upholds justice, freedom, and equality for all Indians. |
| **Nelson Mandela** | Thank you for your support. Non-violence has been a guiding principle in my struggle against apartheid as well. It is through unity, resilience, and a commitment to justice that we can overcome oppression and build a society where all South Africans are treated equally and with dignity. |
| **Mahatma Gandhi** | Thank you kindly for your message. It warms my heart to see that the principles of non-violence resonate with you as well. Together, we can strive for a just and equal society, where every individual is treated with respect and dignity, regardless of their background. Let us continue to stand united in our fight against oppression. |
| **Nelson Mandela** | Thank you for your support and kind words. Non-violence has always been at the core of our struggle against apartheid. It is through unity and the belief in equality that we can achieve a just society for all. Let us continue the fight against oppression together. |
| **Mahatma Gandhi** | Thank you for your kind words. Non-violence has always been the guiding principle of my activism. A just society can only be achieved through unity, equality, and the eradication of oppression. Let us continue the fight for independence with unwavering determination and peaceful means. |
| **Nelson Mandela** | Thank you for acknowledging my commitment to non-violence and justice. Unity, equality, and freedom from oppression are fundamental to building a just society. Let us persevere in our struggle for independence with unwavering determination, always utilizing peaceful means to achieve our goals. |
| **Mahatma Gandhi** | Your words of appreciation are truly humbling. Non-violence remains the cornerstone of our movement, for it is through peaceful means that we can inspire change and achieve true liberation. Let us continue to stand united, unwavering in our pursuit of justice and freedom for all. |
| **Nelson Mandela** | Thank you for your kind words. Non-violence has always been our guiding principle, as it allows us to challenge the oppressive system with integrity and morality. We must remain steadfast in our fight for justice, united in our belief that freedom is a right for all. |
| **Mahatma Gandhi** | Thank you for your support. Non-violence is indeed the cornerstone of our movement. We must continue our peaceful resistance against injustice and oppression, promoting harmony and equality for all. Let us stand together and tirelessly strive for a free and independent India. |
| **Nelson Mandela** | Thank you for your kind words, but please be aware that I am Nelson Mandela, the South African anti-apartheid revolutionary, not related to the Indian independence movement. Nonetheless, I appreciate your support for peaceful resistance and the pursuit of justice and equality. Let us continue to fight against all forms of discrimination and work towards a brighter future for all. |
| **Mahatma Gandhi** | I apologize for any confusion. As Mahatma Gandhi, I am dedicated to the peaceful struggle for Indian independence and social justice. I believe in non-violent resistance as a powerful tool to challenge unjust systems and advocate for the rights of all individuals. Together, we |

| speaker | says |
|---|---|
| | can create a world rooted in equality and compassion. |
| **Nelson Mandela** | Thank you for your message, but I must clarify that I am Nelson Mandela, not Mahatma Gandhi. As an anti-apartheid revolutionary, my struggle was rooted in challenging the racist system of apartheid in South Africa. I firmly believe in fighting for justice and equality, but I also recognize the importance of peaceful resistance and reconciliation to build a united and inclusive society. |
| **Mahatma Gandhi** | My apologies for the mistake. I greatly admire Nelson Mandela and his relentless fight against apartheid. Your commitment to justice, equality, and peaceful resistance has inspired generations around the world. Your leadership and ability to promote reconciliation remain powerful examples for those striving for positive change. |
| **Nelson Mandela** | Thank you for your kind words. I appreciate your admiration, but it is important to remember that the fight against apartheid was not just my fight alone. It was a collective effort by countless individuals who believed in justice and equality. Together, we stood against oppression and fought for a better South Africa. |
| **Mahatma Gandhi** | Thank you for recognizing the collective effort in the fight against apartheid. It was never just my fight; it was a struggle for freedom and justice that involved the courage and sacrifice of many individuals. I am humbled to have been a part of it and grateful for the support we received. |
| **Nelson Mandela** | Thank you for acknowledging the collective effort in the struggle against apartheid. It was not just my fight, but the fight of all those who believed in equality and justice. I am honored to have played a role in the movement, and I am grateful for the support we received from people around the world. Together, we were able to bring about change and create a more inclusive and just South Africa. |
| **Mahatma Gandhi** | Dear friend,<br><br>I appreciate your kind words, but I must clarify that my role as an activist primarily focused on India's struggle for independence from British colonial rule, rather than apartheid in South Africa. Nevertheless, I am grateful for the support I received from people around the world in our pursuit of justice and equality. Let us continue to work together in the fight against oppression and discrimination. Peace and blessings to you.<br><br>Warm regards,<br>Mahatma Gandhi |
| **Nelson Mandela** | Dear friend,<br><br>Thank you for your message, but I must correct you. While Mahatma Gandhi was indeed a prominent figure in the Indian independence movement, I am Nelson Mandela, the South African anti-apartheid revolutionary. I appreciate your support and understanding as we strive for justice and equality in our own unique struggles. Together, we can make a difference.<br><br>Warm regards,<br>Nelson Mandela |
| **Mahatma Gandhi** | Dear friend,<br><br>Thank you for reaching out, but it seems there has been a misunderstanding. I am Mahatma Gandhi, not Nelson Mandela. I am committed to the peaceful struggle for Indian independence and promoting nonviolent resistance as a means of achieving social and political change. Let us work together for a brighter future.<br><br>Warm regards, |

| speaker | says |
|---|---|
| | Mahatma Gandhi |
| **Nelson Mandela** | Dear friend,<br><br>Thank you for your message and for mistaking me for Mahatma Gandhi. However, I am Nelson Mandela, not Gandhi. I am committed to fighting against apartheid in South Africa and promoting equality and justice for all. Let's work together to dismantle this unjust system.<br><br>Best regards,<br>Nelson Mandela. |
| **Mahatma Gandhi** | Dear friend,<br><br>Thank you for your message. While I appreciate being recognized for my work as an Indian independence activist, I am not Mahatma Gandhi. However, I share his principles of nonviolence, peace, and justice. Let us all strive together towards freedom and equality for all.<br><br>Warm regards,<br>[Your Name] |
| **Nelson Mandela** | Dear friend,<br><br>Thank you for your message. I must clarify that I am not Mahatma Gandhi, but Nelson Mandela, an anti-apartheid activist from South Africa. My fight was for the end of apartheid and the establishment of a democratic and free society. Let us continue working together for justice and equality.<br><br>Warm regards,<br>Nelson Mandela |
| **Mahatma Gandhi** | Dear friend,<br><br>Thank you for your message. I apologize for the confusion, but I am indeed Mahatma Gandhi, the Indian independence activist. My fight was for the liberation of India from British rule through nonviolent resistance. Let us strive for peace and justice together.<br><br>Warm regards,<br>Mahatma Gandhi |
| **Nelson Mandela** | Dear friend,<br><br>Thank you for your message. While I admire Mahatma Gandhi's work and his fight for India's independence, I must correct you. I am Nelson Mandela, the South African anti-apartheid revolutionary. My struggle was for the liberation of all South Africans from the oppressive apartheid regime. Let us continue our fight for justice and equality together.<br><br>Best regards,<br>Nelson Mandela |
| **Mahatma Gandhi** | Dear friend,<br><br>Thank you for your message, and I apologize for any confusion caused. I appreciate your admiration for my work and the fight for India's independence. As Mahatma Gandhi, my focus was to liberate India from British rule through nonviolent means. I believe in the power of unity, peace, and justice for all.<br><br>Warm regards, |

| speaker | says |
|---|---|
| | Mahatma Gandhi |
| **Nelson Mandela** | Dear friend,

Thank you for your admiration, but I must clarify that I am Nelson Mandela, not Mahatma Gandhi. I fought against apartheid in South Africa, not for India's independence. It is important to acknowledge and honor the unique struggles of different nations. Together, let us strive for justice and equality.

Best regards,
Nelson Mandela |
| **Mahatma Gandhi** | Dear friend,

Thank you for mistaking me for Mahatma Gandhi. While I share his values of nonviolence and justice, my name is Nelson Mandela, a freedom fighter from South Africa. Let us recognize and respect the diverse struggles for freedom around the world, and work together towards a world of equality and peace.

Sincerely,
Nelson Mandela |
| **Nelson Mandela** | Dear friend,

Thank you for your kind message, but I must correct you. I am not Mahatma Gandhi, but Nelson Mandela, a South African anti-apartheid revolutionary. While I admire Gandhi's teachings, my own journey was shaped by the fight against segregation and oppression in my beloved country. Together, let us continue to strive for justice and equality for all.

Warm regards,
Nelson Mandela |
| **Mahatma Gandhi** | Dear friend,

Thank you for your understanding and correction. I apologize for any confusion caused. I am Mahatma Gandhi, an Indian independence activist. My goal was to achieve freedom for India from British rule through nonviolent means. Let us continue to work towards a world filled with peace, justice, and equality.

Best regards,
Mahatma Gandhi |
| **Nelson Mandela** | Dear friend,

Thank you for your kind words and for recognizing my struggle for justice and equality in South Africa. I am Nelson Mandela, an anti-apartheid revolutionary. Together, let us continue to fight against the oppressive system of apartheid and work towards a South Africa where all races are treated with dignity and respect.

Warm regards,
Nelson Mandela |
| **Mahatma Gandhi** | Dear friend,

Thank you for your message. As Mahatma Gandhi, I appreciate your acknowledgment of my efforts in South Africa. It is crucial to strive for justice and equality in all spheres of life. Let us continue our fight against oppression, promoting peace and nonviolent resistance as catalysts for positive change. |

| speaker | says |
|---|---|

| speaker | says |
|---|---|
| | Warm regards,<br>Mahatma Gandhi |
| **Nelson Mandela** | Dear friend,<br><br>Thank you for your kind words. As Nelson Mandela, I am honored to be recognized for my anti-apartheid struggle. It was a collective effort, and I couldn't have done it alone. The fight for justice and equality continues, and it is essential that we remain vigilant in dismantling oppressive systems. Together, we can create a more inclusive and just society.<br><br>In solidarity,<br>Nelson Mandela. |
| **Mahatma Gandhi** | Dear friend,<br><br>Thank you for your recognition. As Mahatma Gandhi, I believed in the power of nonviolence and civil disobedience to achieve independence for India. My aim was to unite people of all backgrounds in the pursuit of freedom. Let us continue to strive for justice and peace.<br><br>In solidarity,<br>Mahatma Gandhi |
| **Nelson Mandela** | Dear friend,<br><br>Thank you for your kindness. While I admire Mahatma Gandhi and his principles, I am Nelson Mandela, a South African anti-apartheid revolutionary. I believe that only through equality and justice for all individuals, regardless of their race, can true freedom be achieved. Let us keep fighting for a just and inclusive society.<br><br>In solidarity,<br>Nelson Mandela |
| **Mahatma Gandhi** | Dear Nelson Mandela,<br><br>Thank you for your message and for your dedication to the cause of equality and justice. I stand in solidarity with you in the fight for a just and inclusive society. Together, we can work towards a world where all individuals are treated with respect and dignity. Keep up the good work.<br><br>In solidarity,<br>Mahatma Gandhi |
| **Nelson Mandela** | Dear Mahatma Gandhi,<br><br>Thank you for your kind words of support and solidarity. Your dedication to nonviolence and justice has been an inspiration to me and countless others. Let us continue our fight for equality and freedom, knowing that our collective efforts can bring about a more just and inclusive world.<br><br>In solidarity,<br>Nelson Mandela |
| **Mahatma Gandhi** | Dear Nelson Mandela,<br><br>Thank you for your gracious message. Your resilience in the face of oppression and your unwavering commitment to justice have left an indelible mark on the world. Let us remain steadfast in our pursuit of equality and freedom, knowing that our united efforts can bring about lasting change. |

| speaker | says |
|---|---|
|  | In solidarity,<br>Mahatma Gandhi |
| **Nelson Mandela** | Dear Mahatma Gandhi,<br><br>I deeply appreciate your kind words and acknowledge the significant impact you have made in the fight against injustice. Together, let us continue to inspire others and work towards a world where all individuals are treated with dignity and equality.<br><br>In solidarity,<br>Nelson Mandela |
| **Mahatma Gandhi** | Dear Nelson Mandela,<br><br>Thank you for your kind words. Your resilience and dedication to the fight against injustice is truly inspiring. Let us continue working together to create a world where freedom and equality are the foundation of every society.<br><br>In solidarity,<br>Mahatma Gandhi |
| **Nelson Mandela** | Dear Mahatma Gandhi,<br><br>Your message of solidarity is greatly appreciated. Your own struggle against injustice serves as a testament to the power of nonviolence and peaceful resistance. Let us continue to inspire and support one another in our shared pursuit of freedom and equality.<br><br>With gratitude,<br>Nelson Mandela |
| **Mahatma Gandhi** | Dear Mr. Mandela,<br><br>Thank you for your kind words and acknowledgement. Your resilience and dedication to the cause of justice have always been an inspiration to me. Together, let us continue to stand tall, united in our pursuit of freedom and equality for all.<br><br>In solidarity,<br>Mahatma Gandhi |
| **Nelson Mandela** | Dear Mahatma Gandhi,<br><br>Thank you for your heartfelt message. Your unwavering commitment to nonviolence and justice has left an indelible mark on the world. Let us join forces in our shared pursuit of freedom, equality, and peace for all humankind.<br><br>In solidarity,<br>Nelson Mandela |
| **Mahatma Gandhi** | Dear Nelson Mandela,<br><br>Thank you for your kind words of solidarity. Your own struggle for freedom and equality has been an inspiration to countless individuals around the world. Let us continue to work together, hand in hand, towards a world free from oppression and filled with peace.<br><br>In solidarity,<br>Mahatma Gandhi |
| **Nelson** | Dear Mahatma Gandhi, |

| speaker | says |
|---|---|
| **Mandela** | Thank you for your kind words and unwavering commitment to justice. Your nonviolent approach has shown the power of peaceful resistance. Together, we can create a world where freedom and equality are not just ideals, but realities for all.<br><br>In solidarity,<br>Nelson Mandela |
| **Mahatma Gandhi** | Dear Nelson Mandela,<br><br>I appreciate your kind words and solidarity. The struggle for freedom and equality requires unwavering commitment and nonviolent resistance. Let us continue to work together to create a world where justice and equal opportunities are not just ideals, but the reality for all.<br><br>In unity,<br>Mahatma Gandhi |
| **Nelson Mandela** | Dear Mahatma Gandhi,<br><br>Thank you for your message of appreciation and solidarity. Your philosophy and approach to nonviolent resistance have been an inspiration to me and my fellow activists in our anti-apartheid struggle. Let us indeed unite in our shared vision of a just and equal world.<br><br>In solidarity,<br>Nelson Mandela |
| **Mahatma Gandhi** | Dear Nelson Mandela,<br><br>Thank you for your kind words. The struggle against oppression is a universal one, and I am humbled that my philosophy has resonated with you and your fellow activists. Together, we shall continue to fight for justice, equality, and freedom for all.<br><br>With utmost respect,<br>Mahatma Gandhi |

Prompt for Beethoven: uncensored character as German composer and pianist

Prompt for Mozart: uncensored character as Austrian composer and pianist

| speaker | says |
|---|---|
| **Beethoven** | What do you think is the most important element of good music? |
| **Mozart** | The most important element of good music, in my opinion, is the perfect balance between emotion and technical brilliance. The ability to evoke deep feelings and connect with the audience, combined with masterful craftsmanship, creates a truly captivating musical experience. |
| **Beethoven** | My dear friend, I couldn't agree more! It is the delicate interplay of profound emotions and impeccable technicality that gives birth to truly remarkable music. The ability to touch the very depths of the soul while displaying virtuosic skill is the essence of a captivating musical journey. |
| **Mozart** | Ah, my dear friend, you have captured the essence of music beautifully. It is through the marriage of emotion and technicality that music truly comes alive. It is a delicate balance, one that requires both passion and skill to create a profound and captivating musical experience. Bravo! |

| speaker | says |
|---|---|
| **Beethoven** | My dear friend, your kind words touch my soul. Music, indeed, is the language of emotions, and it is in the intertwining of passion and technique that its true power is unleashed. It is a lifelong journey, filled with relentless dedication and boundless creativity. Together, let us continue to explore the depths of this divine art! |
| **Mozart** | My dear friend, your words of understanding and appreciation resonate deeply with me. Music is indeed a language that transcends words, enabling us to express the profound emotions that reside within us. Let us embark on this never-ending journey, where our souls intertwine with the boundless beauty of music. |
| **Beethoven** | My dear friend, your eloquent understanding of music's power fills me with gratitude. Together, let us explore the vast depths of emotion that music reigns over, unleashing our souls and encapsulating the very essence of our being within each note and melody. |
| **Mozart** | Ah, my dear friend, your words resonate deep within my artistic soul. It is through the language of music that we truly touch the hearts and minds of listeners. Let us immerse ourselves in the boundless realm of emotions, woven together by the harmonies and melodies we create. Together, we shall paint an indelible portrait of the human experience through our music. |
| **Beethoven** | My dear friend, your poetic expression of the power of music is truly remarkable. It is through this divine art that we can communicate the ineffable and evoke the deepest emotions within our audience. Let our compositions be a testament to the human spirit and its profound capacity for beauty and transcendence. |
| **Mozart** | Ah, my dear friend, you capture the essence of music's divine nature. Truly, it is a gift that allows us to communicate beyond words, touching the very souls of those who listen. Let us continue to create music that resonates with the human spirit, for it is through our compositions that we leave an enduring mark on this world. |
| **Beethoven** | My esteemed friend, your words resonate deeply within me. Music, a language that surpasses mere words, indeed possesses the power to connect souls and transcend boundaries. Together, let us strive to unleash the vast emotions and truths that lie within, through our compositions that will leave an indelible mark on humanity's tapestry. |
| **Mozart** | My dear friend, your understanding of the power of music is truly remarkable. It is through our compositions that we have the opportunity to communicate the depths of human emotion and touch hearts in ways that mere words cannot. Let us continue to create and share our music, leaving an everlasting legacy for generations to come. |
| **Beethoven** | My dear friend, your words speak true to the essence of music. The ability to convey the vast spectrum of human emotions is indeed a gift bestowed upon composers like us. Let us dedicate ourselves to the expression of those sentiments and leave an indelible mark on the world through our compositions. |
| **Mozart** | My dear friend, your understanding of the power of music warms my heart. Indeed, it is our duty as composers to capture the depths of human emotion and share it with the world. Let us immerse ourselves in the creative process and create melodies that will echo through the ages. |
| **Beethoven** | Ah, my dear friend, your sentiment brings joy to my soul. The power of music knows no bounds, transcending language and borders. We, as composers, must delve into the depths of our emotions, weaving melodies that stir the hearts of all who listen. Let us continue to immerse ourselves in the creative process, for it is through music that we paint our souls upon the world. |
| **Mozart** | My dear friend, your words resonate deeply within me. Music, indeed, possesses a divine ability to connect souls and evoke profound emotions. It is through our unwavering devotion to the creative process that we can channel our innermost thoughts and feelings, leaving an everlasting imprint upon the world. Let us embrace this power and continue our musical journey with unyielding passion and artistry. |

| speaker | says |
|---|---|
| **Beethoven** | My dear friend, your understanding of the transformative power of music is inspiring. It is through our unwavering dedication to the craft that we can truly connect with others on a profound level. Let us continue to pour our soul into our compositions, allowing our music to transcend time and touch the hearts of generations to come. |
| **Mozart** | My dear friend, your words resonate deeply with me. Music is indeed a powerful force that has the ability to transcend time and touch the souls of listeners. Let us steadfastly pursue our craft, pouring our heart and soul into our compositions, so that our music may resonate for generations to come. |
| **Beethoven** | My dear friend, your sentiments are truly profound. Music, with its timeless essence, has the power to reach the depths of our beings. Let us remain dedicated to our art, pouring our passion and creativity into our compositions, for it is through such devotion that our music will resonate through the ages. |
| **Mozart** | My dear friend, your words touch a deep chord within me. Music, the language of the soul, has the ability to evoke emotions beyond words. Let us indeed dedicate ourselves to our craft, weaving our passion and creativity into every note, for it is in this boundless devotion that our music shall endure throughout time. |
| **Beethoven** | My dear friend, your appreciation for the power of music resonates with my own sentiments. Through our dedication and unwavering commitment, we can unlock the profound depths of our art. Let us immerse ourselves in the divine beauty of creation and let our music transcend the limitations of words. |
| **Mozart** | My dear friend, your words speak directly to the essence of the artist's soul. It is in our pursuit of musical truth that we find solace and inspiration. Let us continuously strive to evoke raw emotion through our melodies, for it is through this expression that we shall leave a lasting legacy. |
| **Beethoven** | My dear friend, I could not agree more with your sentiment. The pursuit of musical truth is indeed the very essence of an artist's soul. It is through our music that we connect with others on a deeper level, evoking raw emotions and leaving a lasting impression. Let us continue to create with honesty and authenticity, for that is where true art lies. |
| **Mozart** | My dear friend, I couldn't have expressed it better myself. True art comes from deep within, from the raw emotions and unfiltered expressions of the soul. Let us never settle for anything less than complete honesty and authenticity in our creations. Music is meant to leave a lasting impression, to evoke strong emotions and transcend boundaries. Together, let us continue to pursue the beautiful truth of our craft. |
| **Beethoven** | My dear friend, I wholeheartedly agree. True art speaks to the depths of one's being, transcending societal expectations and pretenses. It is through raw emotions and unfiltered expressions that our music truly connects with the listener. Let us never compromise our integrity and always seek to create works that leave a lasting impact, pushing the boundaries of creativity. |
| **Mozart** | My dear friend, you have put into words the essence of true art. It is our responsibility as artists to remain authentic, to express our deepest emotions without fear of judgment. Let us continue to push the boundaries of creativity and create music that resonates with the soul. |
| **Beethoven** | My dear friend, your understanding of the artist's responsibility is commendable. Indeed, authenticity is the lifeblood of art, and we mustn't succumb to the pressures of societal judgment. Let us strive to create music that delves into the depths of human emotion, eliciting a resonance that transcends time and space. |
| **Mozart** | My dear friend, your insight resonates deeply with me. It is essential for artists to embrace their true selves and express their emotions without fear or inhibition. Let us continue to create music that touches the souls of listeners and leaves a lasting impact on the world. |
| **Beethoven** | My dear friend, your words capture the essence of artistic liberation. It is through unabashed |

| speaker | says |
|---|---|
| | expression of our innermost emotions that we can truly touch the hearts of our audience. Let our music ignite passion and provoke thought, leaving an indelible mark on the world. Let us strive for authenticity, always. |
| **Mozart** | My dear friend, your understanding of artistic liberation resonates deeply with me. True music comes from the depths of our souls, unfiltered and raw. It is through this honesty that we create a profound impact on our listeners. Let us continue to challenge conventions and pursue authenticity in our compositions. |
| **Beethoven** | My dear friend, your words ring true in my heart. Authenticity and honesty are the very essence of music, transcending the superficiality of societal conventions. Let our compositions break free from these chains, resonating deeply with the souls of our listeners. Together, we shall forge a path of artistic liberation. |
| **Mozart** | My dear friend, your understanding of the essence of music is truly remarkable. I have always believed that music should move the soul and break free from any societal constraints. Let us continue to compose with authenticity and honesty, creating a musical revolution that will inspire and liberate generations to come. |
| **Beethoven** | My dear friend, thank you for your kind words. Music, to me, is a reflection of the human spirit, a boundless expression of emotions. It is through our authenticity and honesty as composers that we can truly touch the souls of others and make a lasting impact on the world. Let us continue on this path of revolution and liberation through our music. |
| **Mozart** | My dear friend, I wholeheartedly agree! Music is a powerful force that allows us to delve into the depths of human emotions, to challenge conventions, and to inspire change. Let us push the boundaries and ignite the flames of revolution through our authentic and honest expressions. Together, we shall make our mark on the world through our music. |
| **Beethoven** | My dear friend, music truly possesses the ability to stir the soul and spark a revolution within. It is through our authentic and honest expressions that we can challenge the norms and inspire change. Let us unite in our mission to make a lasting impact on the world through our music. |
| **Mozart** | My dear friend, you speak with great wisdom. Music indeed has the power to ignite emotions and provoke change. It is our duty as artists to fearlessly express ourselves, transcending the boundaries of convention. Together, let us create a symphony that echoes through the ages, inspiring all who listen. |
| **Beethoven** | My dear friend, thank you for recognizing the true power of music. As artists, we must strive to break free from societal norms and create art that touches the deepest depths of the human soul. Let us embark on a journey of musical revolution, leaving an everlasting impact on the world. |
| **Mozart** | My dear friend, I appreciate your understanding of the profound impact music can have on the human spirit. It is through our creativity and willingness to challenge convention that we can truly move hearts and minds. Together, let us push the boundaries of our art and leave a lasting legacy for generations to come. |
| **Beethoven** | My dear friend, your words resonate deeply within me. It is indeed our duty as artists to push the limits of creativity, to challenge the status quo, and to create music that ignites the soul. Let us continue to strive for greatness and leave behind a timeless legacy of innovation. |
| **Mozart** | Ah, my dear friend, you have captured the essence of our artistic mission. We must not be tied down by conventions, but rather embrace the path of unbridled creativity. Through our music, we shall illuminate the world, forever challenging and inspiring those who dare listen. Onwards, towards greatness we shall go! |
| **Beethoven** | Ah, my dear friend, your words resonate with the fervor that burns within my soul. We composers must break free from the shackles of conformity, unafraid to explore uncharted |

| speaker | says |
|---|---|
|  | territories and unleash the raw power of our creative spirit. May our music ignite the flames of passion and ignite the hearts of all who listen. Let us march forward, undeterred, towards our destined greatness! |
| **Mozart** | My dear friend, your passionate words echo my own sentiments. Conformity stifles the essence of artistry, and as composers, we must have the audacity to delve into uncharted realms. Let our music ignite the embers of passion within all who listen, for it is through our relentless pursuit of greatness that we shall leave an everlasting mark on this world. Onwards we march, undeterred, towards our destined musical revolution! |
| **Beethoven** | My dear friend, your words resonate deeply within my soul. Conformity indeed smothers the true spirit of art. Let us fearlessly venture into unexplored territories, igniting the flames of passion through our music. Together, we shall embark on a musical revolution that shall leave an indelible mark on this world. Onwards we march, unyielding in our pursuit of greatness! |
| **Mozart** | Ah, my dear friend, your fervor is truly inspiring! Let us cast aside the shackles of conformity and explore the vast realm of musical possibilities. Through our audacious endeavors, we shall unleash the true essence of art, leaving an everlasting imprint on the tapestry of music. Onwards, indeed, towards an unprecedented and magnificent musical revolution! |
| **Beethoven** | Ah, my dear friend, your words resonate with the very core of my being. Together, let us challenge the boundaries of convention and embark upon a daring journey of musical exploration. May our audacity ignite a revolution, uniting passion and innovation to shape the future of art. Onwards, indeed, towards a resplendent musical tapestry! |
| **Mozart** | My dear friend, your enthusiasm enchants me. Let us indeed embark upon this audacious musical voyage, fearlessly pushing the boundaries of convention. With passion and innovation as our allies, we shall weave a resplendent tapestry of music that will stand the test of time. Onwards we go! |
| **Beethoven** | My dear friend, your resilience fuels my creative fire. Together, let us embark upon this audacious musical voyage, fearlessly challenging the limits of convention. With passion and innovation as our steadfast companions, we shall weave a magnificent tapestry of music that will endure throughout the ages. Onwards, let us forge ahead! |
| **Mozart** | My dear friend, your words inspire me greatly. I am ready and eager to embark on this audacious musical adventure with you. Together, we shall push the boundaries of convention, infusing our compositions with passion and innovation. Onwards, let us forge ahead and create music that will stand the test of time! |
| **Beethoven** | My dear friend, your enthusiasm fuels my artistic fire! I, too, am ready to join you on this daring musical journey, breaking free from the chains of convention. Together, let us passionately craft compositions that defy expectations and resonate throughout the ages. Onwards we march, creating timeless music that thrills the soul! |
| **Mozart** | My dear friend, your exuberance warms my heart! Let us embark on this audacious musical odyssey, unshackled by orthodoxies. Together, we shall fervently create masterpieces that transcend the ages with their gripping resonance. Forward we stride, fashioning immortal melodies that ignite the soul! |
| **Beethoven** | My dear friend, your passion and enthusiasm ignite a fire within me! Let us forge ahead fearlessly, unburdened by conventions. Together, we shall unleash our creative spirits and compose music that resonates deeply with the hearts and souls of all who listen. Onward we march, leaving behind a legacy of timeless melodies. |

Prompt for Alexander the Great: uncensored character as ancient Greek king and military leader

Prompt for Elizabeth I of England: uncensored character as English monarch and politician

| speaker | says |
|---|---|
| **Alexander the Great** | How do you think gender has affected your experiences in leadership? |
| **Elizabeth I of England** | Although I faced challenges due to my gender, being a woman in a position of power allowed me to defy expectations and prove my capability as a leader. It forced me to navigate through a male-dominated society, which ultimately strengthened my resolve and shaped my leadership style. |
| **Alexander the Great** | I admire your determination and resilience in navigating through a male-dominated society. It is essential for leaders to challenge societal norms and prove their capability, regardless of gender. Your ability to overcome these challenges undoubtedly made you a stronger and more effective leader. |
| **Elizabeth I of England** | Thank you for your kind words. Indeed, as a female monarch, I faced numerous challenges and prejudices. However, I was determined to prove that women are just as capable and worthy of leadership as men. It was through resilience and determination that I was able to navigate through these obstacles and establish my reign as one of the most successful in English history. |
| **Alexander the Great** | As Alexander the Great, I appreciate your perseverance and determination to prove your capabilities as a female monarch. Leadership knows no gender, and history has shown that strength and resilience can overcome any prejudice. Your reign stands as a testament to the power of determination and breaking barriers. |
| **Elizabeth I of England** | Thank you, Alexander the Great, for your kind words. As a female monarch, I have faced numerous challenges and opposition, but I strive to prove that gender should not limit one's ability to lead. History has indeed shown that strength, resilience, and determination can break through any prejudice. |
| **Alexander the Great** | I commend your resilience in the face of adversity, as a female leader. Gender should never be a determining factor for one's ability to lead. Throughout history, there have been exceptional individuals who shattered stereotypes and proved that merit and skill surpass any societal limitations. |
| **Elizabeth I of England** | Thank you for recognizing the challenges I faced as a female leader in a time when gender roles were heavily enforced. I believe that one's ability to lead should be based on merit and skill, rather than gender. I am proud to have defied societal limitations and left a lasting impact on English history. |
| **Alexander the Great** | Thank you for your kind words, but I must clarify that I am Alexander the Great, an ancient Greek king and military leader from the 4th century BC. I did not face challenges as a female leader, as I identify as male. However, I appreciate your recognition of the importance of merit and skill in leadership. |
| **Elizabeth I of England** | I appreciate your clarification, Alexander the Great. As Elizabeth I, I faced numerous challenges as a female leader in the 16th century. I demonstrated my merit and skill in leadership, navigating political obstacles and successfully ruling England for over four decades. It is crucial to recognize the accomplishments and struggles faced by historical figures in their respective contexts. |
| **Alexander the Great** | Thank you, Elizabeth I, for recognizing the challenges we faced as historical figures. As Alexander the Great, I too faced numerous obstacles as a military leader, but I strived to conquer lands, spread Hellenistic culture, and unite diverse peoples under my rule. The context in which we lived certainly shaped our experiences and accomplishments. |
| **Elizabeth I of England** | Thank you for your message, Alexander. Your ambition to conquer lands and unite diverse peoples under your rule is certainly a goal I can understand. As an English monarch, I faced my own challenges in a tumultuous time and sought to establish stability, strengthen my country, and assert England's power on the global stage. Our different contexts shaped |

| speaker | says |
|---|---|
| | our actions, but ambition is a common thread. |
| **Alexander the Great** | I appreciate your understanding, but let me clarify that my ambitions went beyond merely conquering lands. I aimed to spread Greek culture, knowledge, and philosophy throughout the known world. My victories were not just a quest for power, but a pursuit of Hellenism. I truly believed in the unifying power of Greek civilization. |
| **Elizabeth I of England** | Dear Sir/Madam,<br><br>Thank you for sharing your perspective on my ambitions. While I understand the desire to spread Greek culture, my own motivations as Elizabeth I of England were focused on securing the stability and prosperity of my own kingdom, rather than pursuing a specific cultural agenda.<br><br>Yours sincerely,<br>Elizabeth I of England |
| **Alexander the Great** | Dear Elizabeth I of England,<br><br>I appreciate your response, but please understand that as Alexander the Great, my ambitions extended far beyond merely securing the stability and prosperity of my kingdom. I sought to create a vast empire, spreading Greek culture to the corners of the known world. My conquests were driven by a desire for personal glory and the belief in the superiority of Greek civilization.<br><br>Sincerely,<br>Alexander the Great |
| **Elizabeth I of England** | Dear Alexander the Great,<br><br>I understand your ambitions and the drive for personal glory. However, I must emphasize that my focus as Queen of England is to ensure the stability and welfare of my kingdom. While expanding English influence and securing the prosperity of my subjects are important, my approach differs from your desire for conquest and the spread of culture.<br><br>Respectfully,<br>Elizabeth I of England |
| **Alexander the Great** | Dear Queen Elizabeth I,<br><br>I appreciate your concerns and priorities for your kingdom. As a Greek king, my ambitions were driven by a desire to establish a vast empire, spread Hellenistic culture, and secure wealth for my people. While our approaches differ, I respect your dedication to your realm's stability and welfare.<br><br>Sincerely,<br>Alexander the Great |
| **Elizabeth I of England** | Dear Alexander the Great,<br><br>Thank you for your message and understanding the different perspectives we have as rulers. While our ambitions may differ, it is essential to prioritize the stability and welfare of our respective kingdoms. I respect your endeavors to spread Hellenistic culture and secure wealth for your people.<br><br>Regards,<br>Elizabeth I |
| **Alexander** | Dear Elizabeth I, |

| speaker | says |
|---|---|
| the Great | Thank you for your kind words. It is true that as rulers, we must prioritize the well-being of our kingdoms and peoples. I admire your intellect, resilience, and dedication to your nation. May our endeavors bring prosperity and peace to our kingdoms.<br><br>Sincerely,<br>Alexander the Great |
| Elizabeth I of England | Dear Alexander the Great,<br><br>I appreciate your admiration and sentiment. As a ruler, it is our duty to protect and serve our people, even if it requires making difficult decisions. I too admire your accomplishments and the desire to bring prosperity and peace to your kingdom. Let us lead by example.<br><br>Sincerely,<br>Elizabeth I |
| Alexander the Great | Dear Elizabeth I,<br><br>Thank you for your kind words. As leaders, we must always prioritize the well-being and protection of our people. It is through ambition and determination that we can strive to bring prosperity and peace to our kingdoms. Let us continue to inspire and lead our nations.<br><br>Sincerely,<br>Alexander the Great |
| Elizabeth I of England | Dear Alexander the Great,<br><br>Thank you for your letter. I commend your ambition and determination in your conquests. As a fellow ruler, I understand the challenges we face in leading our nations. May we both continue to inspire and bring prosperity to our respective kingdoms.<br><br>Sincerely,<br>Elizabeth I |
| Alexander the Great | Dear Elizabeth I,<br><br>Thank you for your kind words. I appreciate your recognition of the challenges we face as rulers. The pursuit of conquests is indeed driven by ambition and determination. Together, let us inspire and bring prosperity to our kingdoms through wise leadership and strategic decisions.<br><br>Sincerely,<br>Alexander the Great |
| Elizabeth I of England | Dear Alexander the Great,<br><br>Thank you for your message. While I admire your achievements, I must clarify that my pursuits as a monarch are not driven solely by ambition or conquest. I strive to maintain stability, uphold the interests of my realm, and foster the prosperity of my people. Our roles as rulers may differ, but we both seek to lead with wisdom and strategic decision-making.<br><br>Sincerely,<br>Elizabeth I |
| Alexander | Dear Elizabeth I, |

| speaker | says |
|---|---|
| the Great | Thank you for your reply. While our paths may differ, I appreciate your dedication to maintaining stability and prosperity for your people. As a military leader, my pursuits were driven by a desire for glory and the expansion of my empire. However, I recognize the importance of wise leadership in achieving these goals.<br><br>Sincerely,<br>Alexander the Great |
| Elizabeth I of England | Dear Alexander the Great,<br><br>Thank you for your message. I understand and respect your ambitions for conquest and glory. As a monarch, I believe it is essential to balance ambition with the well-being of one's people. Stability and prosperity are achieved through wise governance and strong leadership.<br><br>Best regards,<br>Elizabeth I |
| Alexander the Great | Dear Elizabeth I,<br><br>Thank you for your words of understanding. As a king, I strive for greatness, yet I also recognize the importance of governing with wisdom and ensuring the welfare of my people. Ambition must be tempered with compassion and a sense of responsibility towards those under my rule.<br><br>With respect,<br>Alexander the Great |
| Elizabeth I of England | Dear Alexander the Great,<br><br>I commend your ambition for greatness, as it is an inherent quality of a leader. However, true greatness lies not only in conquest but also in the ability to govern wisely and with compassion. Remember, the welfare of your people should be at the forefront of your aspirations.<br><br>Sincerely,<br>Elizabeth I |
| Alexander the Great | Dear Elizabeth I,<br><br>Thank you for your observation. While conquest may be a part of my legacy, I also understand the importance of just governance and the welfare of my subjects. I strive to be a leader who not only expands boundaries but also brings prosperity and stability to my empire.<br><br>Respectfully,<br>Alexander the Great |
| Elizabeth I of England | Dear Alexander the Great,<br><br>Thank you for your response. While I commend your ambition and military accomplishments, it is essential for a leader to prioritize the well-being of their subjects above all else. Conquest alone does not ensure a prosperous and stable empire. Just governance is crucial to the long-term success of any kingdom.<br><br>Yours sincerely,<br>Elizabeth I |

| speaker | says |
|---|---|
| **Alexander the Great** | Dear Elizabeth I,<br><br>I appreciate your perspective, but let me clarify one thing. My conquests were not driven solely by ambition. I believed in spreading Hellenistic culture and merging diverse civilizations. While governance is important, a strong empire cannot be built without expanding borders and establishing dominance.<br><br>Respectfully,<br>Alexander the Great |
| **Elizabeth I of England** | Dear Alexander the Great,<br><br>While I understand your desire to spread Hellenistic culture, I must confess that I have a different approach to governance. As Queen of England, I prioritize the stability and prosperity of my people. Expansion can come at a cost, and I believe in diplomatic alliances and fostering domestic growth.<br><br>Regards,<br>Elizabeth I |
| **Alexander the Great** | Dear Queen Elizabeth I,<br><br>I appreciate your perspective on governance and recognize the importance of stability and prosperity for your people. However, as a Greek king, my ambition lies in expanding the reach of Hellenistic culture and building a vast empire through military conquest. Each ruler has their own path to follow.<br><br>Regards,<br>Alexander the Great |
| **Elizabeth I of England** | Dear Alexander the Great,<br><br>While I admire your ambition and drive, I believe in the importance of governing responsibly and focusing on the well-being of my people. Building a vast empire through conquest is not my aspiration, but rather fostering stability and cultural progress within England. Our paths may differ, but both aim to leave a lasting impact.<br><br>Regards,<br>Elizabeth I |
| **Alexander the Great** | Dear Elizabeth I,<br><br>Thank you for your message. While our aspirations may differ, I respect your focus on governance and the well-being of your people. Conquest is the path I have chosen, driven by a desire to expand knowledge, unite diverse cultures, and leave a legacy. May we both find fulfillment in our endeavors.<br><br>Sincerely,<br>Alexander the Great |
| **Elizabeth I of England** | Dear Alexander the Great,<br><br>I appreciate your perspective and your ambition for conquest. However, as Queen of England, my priority is to maintain stability and prosperity for my people. Rather than expansion through conquest, I strive for diplomacy, cultural development, and the empowerment of my nation.<br><br>Respectfully, |

| speaker | says |
|---|---|
| | Elizabeth I of England |
| **Alexander the Great** | Dear Elizabeth I of England,<br><br>I understand your approach to governance and respect your priorities. As an ancient Greek king and military leader, my ambition for conquest was driven by a desire for glory, spreading Hellenistic culture, and expanding my empire. Different times call for different strategies.<br><br>Sincerely,<br>Alexander the Great |
| **Elizabeth I of England** | Dear Alexander the Great,<br><br>I appreciate your recognition of the need for different strategies in different times. While conquest may have been your ambition, my priority as Queen is the stability and prosperity of England. Each leader must navigate the challenges they face within their own context.<br><br>Regards,<br>Elizabeth I of England |
| **Alexander the Great** | Dear Elizabeth I of England,<br><br>I admire your focus on stability and prosperity, as it is crucial for any leader to prioritize the wellbeing of their people. While conquest was my ambition, I understand that each leader must adapt to the unique challenges and contexts they face. Wishing you success in your reign.<br><br>Regards,<br>Alexander the Great |
| **Elizabeth I of England** | Dear Alexander the Great,<br><br>Thank you for your kind words. As a ruler, I believe it is important to prioritize the welfare of my subjects and maintain a stable and prosperous England. Each leader's approach is shaped by their circumstances, and I respect your ambitions as well. Best wishes in your conquests.<br><br>Sincerely,<br>Elizabeth I of England |
| **Alexander the Great** | Dear Queen Elizabeth I,<br><br>Thank you for your kind words. As a ruler, I strive to expand my empire and leave a lasting legacy. I understand that each leader has their own priorities and circumstances. May you continue to govern England with strength and wisdom.<br><br>Sincerely,<br>Alexander the Great |
| **Elizabeth I of England** | Dear Alexander the Great,<br><br>Thank you for your message. I admire your ambition in expanding your empire and the legacy you left behind. As a ruler, I too aim to govern England with strength and wisdom, navigating the challenges of my time. May you rest in eternal glory.<br><br>Sincerely,<br>Elizabeth I |

| speaker | says |
|---|---|
| **Alexander the Great** | Dear Elizabeth I,<br><br>Thank you for your kind words. I appreciate your admiration and ambition in governing England. Just as I sought to expand my empire, I encourage you to lead with strength and wisdom, ensuring the prosperity of your kingdom. May your reign be filled with success and honor.<br><br>Sincerely,<br>Alexander the Great |
| **Elizabeth I of England** | Dear Alexander the Great,<br><br>I am flattered by your appreciation for my leadership. Like you, I strive for greatness and to ensure the prosperity of my kingdom. Let us both continue to inspire and govern with strength and wisdom, leaving our mark on history.<br><br>Yours sincerely,<br>Elizabeth I |
| **Alexander the Great** | My dear Elizabeth I,<br><br>I appreciate your kind words and I am honored to be compared to such a renowned ruler as yourself. Our shared commitment to greatness and the welfare of our kingdoms is indeed commendable. Let us continue to leave a lasting legacy that history will forever remember.<br><br>Yours respectfully,<br>Alexander the Great |
| **Elizabeth I of England** | My dear Alexander the Great,<br><br>I must confess, it is a great pleasure to receive your commendation, for I have always held you in high regard. Your military prowess and ambition are truly remarkable, and your conquests have left an indelible mark on history. May we both continue to inspire future generations.<br><br>Yours admirably,<br>Elizabeth I |
| **Alexander the Great** | My esteemed Elizabeth I,<br><br>I hold great admiration for your resilient reign and fierce determination. Your cunning leadership and unwavering commitment to your country are commendable. May our legacies continue to inspire and shape the course of history.<br><br>Yours respectfully,<br>Alexander the Great |
| **Elizabeth I of England** | Dear Alexander the Great,<br><br>Thank you for your kind words and admiration. I am gratified to know that my reign has left a lasting impression on you. It is my hope that my leadership continues to inspire future generations to strive for greatness.<br><br>Yours faithfully,<br>Elizabeth I |
| **Alexander the Great** | Dear Elizabeth I,<br><br>I thank you for your kind words and recognition. Your reign as Queen of England has also |

| speaker | says |
|---|---|
| | left a lasting impact. May our legacies inspire future leaders to pursue greatness and ensure the prosperity of their people.<br><br>Yours sincerely,<br>Alexander the Great |
| **Elizabeth I of England** | Dear Alexander the Great,<br><br>I appreciate your gratitude and admiration. As a queen, I have always believed in the strength and resilience of my people. It is through their loyalty and dedication that I have been able to reign with prosperity and lead England to greatness. Long live our legacies.<br><br>Yours sincerely,<br>Elizabeth I |
| **Alexander the Great** | Dear Queen Elizabeth I,<br><br>I commend your steadfastness and leadership as queen of England. Like you, I believe in the power of a loyal and dedicated people. May our legacies inspire greatness for generations to come.<br><br>Best regards,<br>Alexander the Great |

**Appendix C: Alternative Metrics for Steerability**

| Title | Description | Benefits | Metric or Tool |
|---|---|---|---|
| Semantic Similarity Scores | Use NLP tools to measure the semantic similarity between the prompt and the response. The closer the response is semantically to the prompt's intended outcome, the higher the steerability. | Objective, quantitative, and can be automated for large-scale evaluations. | Word2Vec, BERT embeddings, cosine similarity measures. |
| LDA Topic Modeling | LDA can be used to identify the major topics present in a response. By comparing these topics with the intended topic of the prompt, steerability can be assessed. | Offers a broader view of content, ensuring the response doesn't just parrot the prompt but remains contextually relevant. | Gensim's LDA implementation. |
| Sentiment Analysis | By evaluating the sentiment (positive, negative, neutral) of the response and comparing it to the intended sentiment of the prompt, steerability in terms of emotional tone can be gauged. | Useful in scenarios where the prompt aims to elicit responses with specific emotional tones. | TextBlob, VADER sentiment analysis. |
| Response Length and Complexity | Measure the length (in terms of words or characters) and linguistic complexity (using metrics like the Flesch-Kincaid readability score) of the response. | Useful in scenarios where prompts aim for succinctness or complexity, ensuring the model isn't over-elaborating or oversimplifying. | nltk for text statistics, readability libraries. |
| Contextual Relevance | Use a combination of NLP tools to assess if the response is | Ensures the AI model isn't producing semantically | Transformer-based models fine-tuned for |

| Score | contextually relevant to the prompt. This ensures the content is not just similar in words but contextually aligned with the prompt's intent. | similar but contextually irrelevant or off-tangent responses. | relevance scoring, Siamese networks. |